\documentclass[conference]{IEEEtran}
\IEEEoverridecommandlockouts
\usepackage{bm}
\usepackage{cite}
\usepackage{amsmath,amssymb,amsfonts}
\usepackage{algorithmic}
\usepackage{graphicx}
\usepackage{threeparttable}
\usepackage{textcomp}
\usepackage{algorithmic}
\usepackage{algorithm}
\usepackage{url}
\usepackage{marginnote}  
\usepackage{caption} 
\usepackage{booktabs} 
\usepackage{xcolor} 
\def\BibTeX{{\rm B\kern-.05em{\sc i\kern-.025em b}\kern-.08em
    T\kern-.1667em\lower.7ex\hbox{E}\kern-.125emX}}

\begin{document}

\title{\LARGE \bf{Integrating DeepRL with Robust Low-Level Control in Robotic Manipulators for Non-Repetitive Reaching Tasks*}\\

\thanks{*Funding for this research was provided by the Business Finland partnership project "Future All-Electric Rough Terrain Autonomous Mobile Manipulators" (Grant No. 2334/31/2022).}
}

\author{\IEEEauthorblockN{Mehdi Heydari Shahna, Seyed Adel Alizadeh Kolagar, Jouni Mattila}
\IEEEauthorblockA{\textit{Faculty of Engineering and Natural Sciences} \\
\textit{Tampere University}\\
Tampere, Finland \\
mehdi.heydarishahna@tuni.fi}
}

\maketitle

\marginnote{\rotatebox{90}{\textcolor{red}{This paper has been accepted at the \textbf{International Conference on Mechatronics and Automation (ICMA 2024)}, sponsored by the IEEE.}}}  

\begin{abstract}
In robotics, contemporary strategies are learning-based, characterized by a complex black-box nature and a lack of interpretability, which may pose challenges in ensuring stability and safety. To address these issues, we propose integrating a collision-free trajectory planner based on deep reinforcement learning (DRL) with a novel auto-tuning low-level control strategy, all while actively engaging in the learning phase through interactions with the environment. This approach circumvents the control performance and complexities associated with computations while addressing nonrepetitive reaching tasks in the presence of obstacles. First, a model-free DRL agent is employed to plan velocity-bounded motion for a manipulator with 'n' degrees of freedom (DoF), ensuring collision avoidance for the end-effector through joint-level reasoning. The generated reference motion is then input into a robust subsystem-based adaptive controller, which produces the necessary torques, while the cuckoo search optimization (CSO) algorithm enhances control gains to minimize the stabilization and tracking error in the steady state. This approach guarantees robustness and uniform exponential convergence in an unfamiliar environment, despite the presence of uncertainties and disturbances.
Theoretical assertions are validated through the presentation of simulation outcomes.
\end{abstract}


\section{INTRODUCTION}
Robot control involves three levels of abstraction: motor, motion, and task levels \cite{liu2022robot}.
For instance, tasks involving the tracking of predefined joint trajectories primarily fall under motor-level control. However, a more comprehensive approach is necessary for goal-reaching tasks, one involving both motion planning and motor control. Finally, complex tasks, such as welding and polishing require the integration of all three control levels.
Autonomous robot systems rely heavily on motion planning to tackle the challenge of determining a viable, seamless, and collision-free route within a robot's configuration space from an initial to a target point \cite{elbanhawi2014sampling, kroemer2021review}. This identified path can then be followed by a low-level controller \cite{mueller2019modern}. Motion planning commonly employs either sampling-based approaches \cite{karaman2011sampling} or optimization-based methods \cite{mukadam2018continuous}. The former of which guarantees a globally optimum infinite compute time, demonstrating completeness \cite{karaman2011sampling}. However, in practice, they often exhibit sample inefficiency and have the propensity to generate non-smooth trajectories \cite{hauser2010fast}. Conversely, optimization-based planners refine initial trajectories through preconditioned gradient descent \cite{mukadam2018continuous} or stochastic update rules \cite{urain2022learning}, allowing the incorporation of such desired properties as smoothness into optimized costs. However, planners based on optimization rely on proficient initialization and may become ensnared in local minima owing to the non-convex characteristics of intricate problems.
As a current focal point of interest, learning-based methods have gained widespread use for addressing intricate challenges in path and trajectory planning \cite{carvalho2023motion}.
Specifically, reinforcement learning (RL) stands out as an exemplary machine learning paradigm, as its algorithms inherently acquire knowledge through direct interactions with the environment \cite{tamizi2023review}.
As shown in Fig. 1, in the RL, an agent engages with the environment, encounters diverse situations, and receives rewards. By harnessing this experiential knowledge, agents can acquire the ability to make the best decisions in each state \cite{10.5555/3312046}.
Traditional RL is constrained to domains featuring uncomplicated state representations, so to overcome challenges presented by higher-dimensional and more intricate problems, the integration of deep neural networks (DNNs) into RL has given rise to the paradigm known as Deep Reinforcement Learning (DRL) \cite{del2022review}, which has shown promising results in the ﬁeld of robot navigation with continuous action and state space.
DRL algorithms can be classified into two principal types: model-based and model-free methods.
For motion planning, model-free DRL algorithms hold prominence, given the intricacies involved in modeling a dynamic environment \cite{qiu2021applications}. Model-free algorithms encompass policy-based, value-based (DQN-based), and actor–critic methods, the latter combining advantages from both policy and value-based approaches.
Nevertheless, DQN-based methods face limitations in addressing problems characterized by discrete and lower-dimensional action spaces as well as deterministic policies. Simultaneously, policy gradient approaches demonstrate versatility by accommodating continuous action spaces and the representation of stochastic policies. A sample-efficient strategy involves employing an actor–critic architecture capable of effectively utilizing off-policy data \cite{liu2021deep}.
Because Soft Actor–Critic-based (SAC) algorithms use an entropy term in their objective function, they can find the optimal solution to the high-dimensional problem, suggesting that the SAC-based algorithm outperformed the existing results\cite{prianto2021deep}. Recent trends in robotics favor embedding established control principles, such as PD-controlled joint positions \cite{rudin2022learning}, or impedance control \cite{martin2019variable} into the action space, diverging from earlier end-to-end policies that directly output low-level control commands, such as joint torques \cite{wahlstrom2015pixels, lillicrap2015continuous}.
This shift simplifies learning by outputting higher-level policy control commands, such as reference joint velocities, which are subsequently processed by low-level controllers. However, learning policies concerning this joint torque action space can be complex, as they need to grasp the intricacies of the robot's kinematics and dynamics, especially because rewards are typically expressed using task space properties \cite{aljalbout2023role}.
In addition, despite considerable advancements in learning-based joint-level control, it seldom achieves exponentially fast convergence to the reference trajectories, a factor prominent in ensuring control stability and fast goal achievement. This form of stability is capable enhance the interpretability of overall robot motions and generalizes learned policies to handle a broader range of situations and variations.
Furthermore, achieving optimal results through low-level control may necessitate meticulous parameter tuning. Striking the proper balance between performance aspects can be challenging in unknown environments \cite{9161291}, especially considering that the overall performance of the control structure depends on the low-level control's ability to respond quickly and effectively to the learning-based motion planner.
Metaheuristic algorithms, often inspired by nature, have become widely adopted tools for optimization due to their straightforward application and low computational demands, currently among the most extensively used algorithms \cite{yang2013cuckoo}. Unlike Genetic Algorithms (GAs), cuckoo search optimization (CSO) stands out with a single specific parameter, streamlining the focus on general parameter selection. This distinctive feature makes CSO particularly suitable for perfecting control gains. As such, investigations indicated that CSO exhibits the capacity to demonstrate considerably greater efficiency than Particle Swarm Optimization, GAs, and Artificial Bee Colony \cite{yang2009cuckoo, abdelaziz2015cuckoo, bingul2018novel}. To enhance the control performance of goal-reaching tasks for a manipulator with 'n' degrees of freedom (DoF), this paper introduces a novel, robust, joint-level control strategy to perform nonrepetitive tasks by utilizing a velocity-bounded, model-free DRL agent for obstacle-avoiding motion planning. We train the DRL agent in conjunction with a low-level robust control section within the closed-loop system. Simultaneously, optimal gains are adjusted using CSO tailored to minimize time-domain criteria. In summary, this study makes notable contributions to the field of robot control, as follows: (1) all components of the framework, involving the integration of a collision-free DRL-based trajectory planner with a joint-level robust control strategy as well as tuning parameters through CSO, actively participate in the learning phase through interactions with the unknown environment. (2) This approach significantly reduces computational demands compared to methods relying solely on continuous DRL control action. Such methods necessitate handling both the kinematics and dynamics of the robot \cite{aljalbout2023role}. (3) This robust control approach compensates for potential uncertainties and disturbances arising from both the unknown environment and the integration of the DRL agent with the low-level controller.
(4) It ensures uniformly exponential stability of the n-DoF manipulator mechanism. The subsequent sections of the article are structured in the following manner:
Section II presents the modeling of the n-DoF manipulator system. Section III introduces the DRL motion planning architecture. Section IV presents the robust low-level control design by utilizing an adaptive subsystem-based strategy, optimizing the controller gains by customizing CSO, as well as presents the stability analysis.
In Section V, the proposed control strategy is validated by developing the manipulator system provided in \cite{humaloja2021decentralized} in consideration of obstacles in nonrepetitive tasks.
\vspace{-0.3cm}
\begin{figure}[h] 
    \centering
     \scalebox{0.8}{\includegraphics[trim={0cm 0.0cm 0.0cm 0cm},clip,width=\columnwidth]{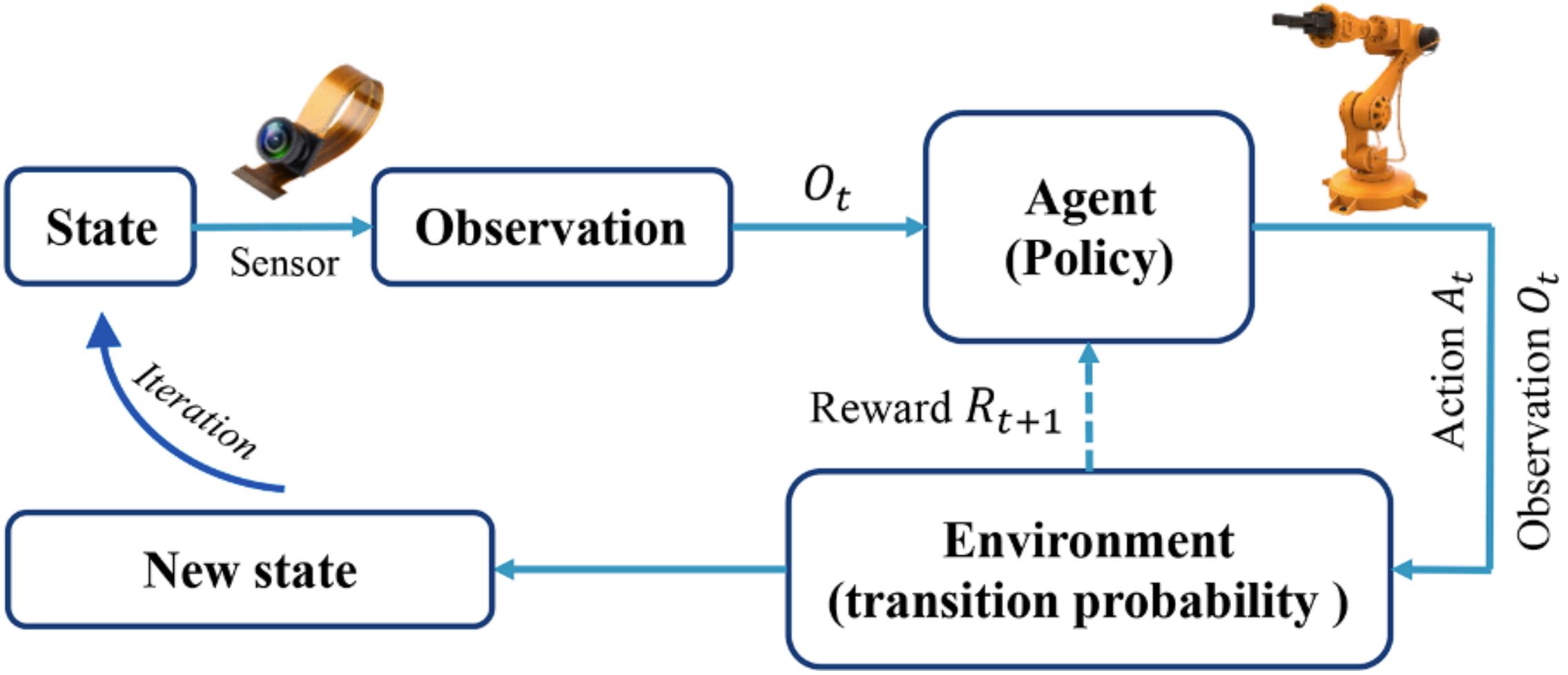}}
    \caption{The architecture of RL-based learning \cite{liu2022robot}.}
    \label{fig33}
\end{figure}

\section{Modeling an n-DoF Robotic Manipulator System}
The dynamic model for an n-DoF robotic manipulator is provided, as follows \cite{nohooji2020constrained}:
\begin{equation}
\small
\begin{aligned}
\label{equation: 1}
{M}({q}) {\ddot{q}}= &{\tau}-{C_m}({q}, {\dot{q}}) {\dot{q}}-{f}({\dot{q}})-{G}({q})-{{\tau}_L}
\end{aligned}
\end{equation}
where ${q} \in \mathbb{R}^n$ represents the joint angles. The mapping ${M(q)}: \mathbb{R}^n \rightarrow \mathbb{R}^{n\times n}$ describes the inertia properties, and ${C_m(q, \dot{q})} : \mathbb{R}^n \times \mathbb{R}^n \rightarrow \mathbb{R}^{n \times n}$ accounts for the centrifugal and Coriolis forces. The function ${G(q)}: \mathbb{R}^n \rightarrow \mathbb{R}^n$ computes the gravitational torques, whereas the function ${f(\dot{q})}: \mathbb{R}^n \rightarrow \mathbb{R}^n$ represents the resistance to movement. The vector ${\tau} \in \mathbb{R}^n$ includes torques applied at the joints, and ${\tau}_L \in \mathbb{R}^n$ characterizes external disturbances affecting the joints. To simplify the manipulator dynamics in the subsystem-based form, we transform \eqref{equation: 1}, as follows:
\begin{equation}
\small
\begin{aligned}
\label{equation: 2}
{\dot{x}_1}(t) = {x_2}(t), \hspace{0.1cm}\dot{x}_2(t) ={M^{-1}}({x_1}) {\tau}+{F}({x_1},{x_2})+ {\tau}_d (t)\\
\end{aligned}
\end{equation}
where ${x}_1={q}=[x_1(1),\ldots,x_1(n)]^\top$, and ${x}_2={\dot{q}}=[x_2(1),\ldots,x_2(n)]^\top$ are the system states, and ${F}=-{M^{-1}}({C_m} {x_2}+{f}+{G})$ and ${\tau_d}=-{M^{-1}}{\tau_L}$ can be uncertainties and load effects, respectively. In addition, the tracking errors are defined, as follows:
\begin{equation}
\small
\begin{aligned}
\label{equation: 3a}
 {e_1}={x_1}-{x_{1d}}, \hspace{0.2cm}
  {e_2}={x_2}-{x_{2d}}
\end{aligned}
\end{equation}
where ${x_{1d}} \in \mathbb{R}^n$ and ${\dot{x}_{1d}}(={x_{2d}}) \in \mathbb{R}^n$ are the desired position and velocity trajectories, respectively.
\section{Reinforcement Learning-based Trajectory Planning}
The objective of our DRL agent is to plan a motion for a robotic manipulator with 'n' DoFs with joint-level commands, ensuring collision avoidance for the end-effector and constraints on joint velocities. To achieve this goal, we adopted the SAC algorithm (algorithm 1), recognized for its proficiency in solving complex tasks; refer to \cite{haarnoja2018soft,haarnoja2018soft6} for more details. A crucial aspect of the successful learning of the agent involves a precise definition of states, actions, and rewards in the training process.
\subsection{State and Action Representation}
The majority of robotic systems based on RL acquire knowledge in the task space as opposed to the joint space \cite{shao2021concept2robot,martin2019variable,lee2019making}. Operating in the task space, specifically the Cartesian end-effector space, significantly simplifies the learning problem. Nevertheless, these solutions prove inadequate in unstructured environments where collisions between the robot and the surroundings necessitate joint-level reasoning.
Furthermore, the majority of approaches rely on solving inverse kinematics or dynamics equations to translate the task space into joint space, primarily for collision avoidance \cite{kumar2021joint}.
The agent can output desired positions, velocities, accelerations, or torques, each of them has its pros and cons. Here, we select the joint velocities as our actions due to their high performance in achieving tasks \cite{aljalbout2023role}. As such, we can constrain the velocities for each joint according to the motor properties or tip velocity. 
In this paper, for goal-reaching tasks, we consider two levels of motion planning. The initial level focuses on reaching the goal without encountering any obstacles. The second level addresses scenarios in which obstacles are present, requiring the end-effector to navigate while avoiding collisions. We selected states, actions, and rewards, drawing inspiration from \cite{kumar2021joint} and \cite{aljalbout2023role}.
In the first situation, the state $s_t$ is defined as:
\begin{equation}
\small
\begin{aligned}
\label{equation: 4a}
s_t=\left\langle p_t, \delta p_t, p, x_1\right\rangle
\end{aligned}
\end{equation}
Here, $p_t=\left(p_{t x}, p_{t y}, p_{t z}\right)$ signifies the Cartesian position of the target. If access to the absolute target position is unavailable, this term can be omitted, although it complicates the training process. $\delta p_t=\left(\delta p_x, \delta p_y, \delta p_z\right)$ represents the end-effector error (distance between the tip and target), $p=\left(p_x, p_y, p_z\right)$ denotes the Cartesian position of the end-effector, and ${x_1} \in \mathbb{R}^n$ represents the angle values for the $\mathrm{n}$ joints of the manipulator.
In the second situation, the state can be adjusted to the following format.
\begin{equation}
\small
\begin{aligned}
\label{equation: 5a}
s_t=<p_t, \delta p_t, p, x_1, \delta p_o>
\end{aligned}
\end{equation}
Here, $\delta p_o=\left(\delta p_{o 1}, \ldots, \delta p_{o m}\right)$ indicates the absolute distance between the end-effector and each obstacle, enabling the manipulator to function in an environment in which the obstacle's absolute position is unknown, necessitating reliance on relative positions obtained through sensors.
In all scenarios, the actions remain consistent, involving joint velocities:
\begin{equation}
\small
\begin{aligned}
\label{equation: 7a}
a_t=<\dot{x}_{1d}>=<x_{2d}>
\end{aligned}
\end{equation}
which input to the low-level control as the desired trajectories that are in predefined finite ranges. These formulations ensure the states are appropriately modified to meet the specified requirements while maintaining consistent actions.
\subsection{Reward Function}
Our reward function $r(s, a)$ is defined as follows:
\begin{equation*}
\scriptsize
\begin{aligned}
\label{equation: 7aaa}
r(s, a) = \begin{cases}
\text{reach\_reward} & \text{if the tip reaches the target} \\
-\text{boundary\_penalty} & \text{if joint angles reach the limits} \\
-\text{obstacle\_penalty} & \text{if there is a collision} \\
-\log_{10}(1 + \text{tip error} - \text{threshold}) & \text{otherwise} \\
(r(s, a) - 1) & (\text{if the tip error increases})
\end{cases}
\end{aligned}
\end{equation*}
The initial component constitutes a positive reward given to the agent upon successful attainment of the target. To ensure controlled movements within the specified workspace, limits are imposed on the manipulator. Should any joint angle exceed the predefined constraints, the agent faces a boundary penalty. In addition, collision occurrences, irrespective of type, result in an obstacle penalty. Alternatively, the reward is determined by a formula, where the tip error represents the absolute distance between the manipulator's tip and the target. The threshold, acting as the acceptable error limit for considering the goal achieved, is factored into this computation. To motivate expedited movements further, an added penalty is introduced. If the tip error increases after progressing to the next step, the agent incurs an additional ``-1" penalty, encouraging the manipulator to accelerate its progress toward the target.
\subsection{Learning Procedure}
To begin the learning process, the environment is initially defined. In the context of our DRL algorithm, both the controller and manipulator model collectively form our environment. This implies the action generated by the agent enters a closed loop, encompassing both the controller and the model, following environment parameters. Subsequently, the workspace is established. This involves outlining the points accessible to the manipulator within the desired region. In addition, the positions of obstacles are specified, and points that are a certain distance from the obstacles are removed, enabling a collision-free workspace. Following the environmental setup, essential variables and functions for the DRL agent are assigned. These include the observation and action space, reset function, step time, and maximum episode time. Simulation resets and the beginning of a new episode occurs when a joint surpasses its limits, the manipulator attains its goal, a collision occurs, or the number of simulation steps exceeds the limit. Subsequently, the SAC agent is created using
the provided information. Training is then initiated by configuring the training parameters accordingly.
\begin{table*}[h!]
\scriptsize
  \centering
  \renewcommand{\arraystretch}{0.3} 
  \colorbox{white!10}{\fbox{%
    \begin{tabular}{p{0.75\linewidth}} 
      \multicolumn{1}{c}{\textbf{Algorithm 1.} Soft Actor-Critic Algorithm \cite{haarnoja2018soft,haarnoja2018soft6}.} \\
      \\
      \hspace{0.4cm}\textbf{Input}: Initial parameters ${\theta}$ (Q-function parameters) , ${\psi}$ (Value network wights), and $\phi$ (Policy network weights).  \\
      \hspace{0.4cm}\textbf{Output}: Optimized parameters ${\theta}$, ${\psi}$, and $\phi$.\\
      \\
      {\small 1}\hspace{0.7cm}Initialize target network weights $\bar{\theta} \leftarrow \theta$, $\bar{\psi} \leftarrow \psi$;\\
      {\small 2}\hspace{0.7cm}Initialize an empty replay pool $\mathcal{D} \leftarrow \emptyset$;\\
      {\small 3}\hspace{0.7cm}\textbf{for each iteration do}\\
      {\small 4}\hspace{1.2cm} \text{Set the manipulator position to the defined initial state};\\
      {\small 5}\hspace{1.2cm} \text{Randomly choose the target position from the collision-free workspace}; \\   
      {\small 6}\hspace{1.2cm} \textbf{for each environment step do}\\
      {\small 7}\hspace{1.7cm} Sample action from the policy $\mathbf{a}_t \sim \pi_\phi\left(\mathbf{a}_t \mid \mathbf{s}_t\right)$;\\
      {\small 8}\hspace{1.7cm} Sample transition from the environment $\mathbf{s}_{t+1} \sim p\left(\mathbf{s}_{t+1} \mid \mathbf{s}_t, \mathbf{a}_t\right)$; \\
      \hspace{2.5cm} (including giving action ($a_t=x_{2d}$) as input to \textbf{algorithm 3} and waiting for the environment response) \\
      {\small 9}\hspace{1.7cm} Store the transition in the replay pool $\mathcal{D} \leftarrow \mathcal{D} \cup\left\{\left(\mathbf{s}_t, \mathbf{a}_t, r\left(\mathbf{s}_t, \mathbf{a}_t\right), \mathbf{s}_{t+1}\right)\right\}$; \\
      {\small 10}\hspace{1.1cm} \textbf{end} \\      
      {\small 11}\hspace{1.2cm}\textbf{for each gradient step do}\\
      {\small 12}\hspace{1.5cm} Update value network weights $\psi \leftarrow \psi-\lambda_v \hat{\nabla}_\psi J_v(\psi)$; \\
       \hspace{2.5cm} $J_V(\psi)=\mathbb{E}_{\mathbf{s}_t \sim \mathcal{D}}\left[\frac{1}{2}\left(V_\psi\left(\mathbf{s}_t\right)-\mathbb{E}_{\mathbf{a}_t \sim \pi_\phi}\left[Q_\theta\left(\mathbf{s}_t, \mathbf{a}_t\right)-\log \pi_\phi\left(\mathbf{a}_t \mid \mathbf{s}_t\right)\right]\right)^2\right]$; \\
      {\small 13}\hspace{1.5cm} Update the Q-function parameters $\theta_\xi \leftarrow \theta_i-\lambda_Q \hat{\nabla}_{\theta_\xi} J_Q\left(\theta_\xi\right)$ for $\xi \in\{1,2\}$;\\
     \hspace{2.5cm} $J_Q(\theta)=\mathbb{E}_{\left(\mathbf{s}_t, \mathbf{a}_t\right) \sim \mathcal{D}}\left[\frac{1}{2}\left(Q_\theta\left(\mathbf{s}_t, \mathbf{a}_t\right)-r\left(\mathbf{s}_t, \mathbf{a}_t\right)-\gamma \mathbb{E}_{\mathbf{s}_{t+1} \sim p}\left[V_{\bar{\psi}}\left(\mathbf{s}_{t+1}\right)\right]\right)^2\right]$; \\
      {\small 14}\hspace{1.5cm} Update policy weights $\phi \leftarrow \phi-\lambda_\pi \hat{\nabla}_\phi J_\pi(\phi)$; \\
      \hspace{2.5cm} $J_\pi(\phi)=\mathbb{E}_{\mathbf{s}_t \sim \mathcal{D}, \epsilon_t \sim \mathcal{N}}\left[\log \pi_\phi\left(f_\phi\left(\epsilon_t ; \mathbf{s}_t\right) \mid \mathbf{s}_t\right)-Q_\theta\left(\mathbf{s}_t, f_\phi\left(\epsilon_t ; \mathbf{s}_t\right)\right)\right]$; \\
      {\small 15}\hspace{1.5cm} Substitute target network weights $\bar{\psi} \leftarrow \tau \psi+(1-\tau) \bar{\psi}$; \\
      {\small 16}\hspace{1.2cm}\textbf{end} \\
      {\small 17}\hspace{0.6cm}\textbf{end} \\   
    \end{tabular}%
  }}
\end{table*}
\section{ROBUST DRL-BASED CONTROL STRUCTURE}
\subsection{Cuckoo Search Optimization}
The CSO mimics cuckoos' brood parasitism, optimizing by selecting nests with optimal solutions through a random walk with a power-law step-length distribution \cite{yang2013cuckoo}. Similar to \cite{bingul2018novel}, this paper concentrates on minimizing the objective function of a low-level controller that includes time-domain features, such as the overshoot, rise time, settling time, and steady state of the tracking error of the reference trajectories commanded by Section III. The number of cuckoo eggs (control gains) is defined as $\zeta$, deposited in a randomly chosen nest $B_j \in \mathbb{R}^{1 \times {\zeta}}$ for $j=1,...,\eta$ in which $\eta$ is the swarm (population) size.
The best nest $B_{best} \in \mathbb{R}^{1 \times {\zeta}}$, housing optimal solutions, are retained for the next generation of cuckoos, and new nests for cuckoos are generated using a global random walk called Lévy flights with a specific iteration ($=n_{iteration}$), expressed, as follows:
\begin{equation}
\small
\begin{aligned}
\label{equation: 15}
B_{j(new)}=B_j+\ell. \operatorname{levy}(stp)
\end{aligned}
\end{equation}
where $B_{j(new)}$ and $B_j$ are new and current nest, respectively. The parameter $\ell$ is a random number normally distributed, considering the scale of the problem at hand, and the change in position during the flight is considered as follows:
\begin{equation}
\small
\begin{aligned}
\label{equation: 16}
\operatorname{levy}(stp)=0.01 \cdot stp \cdot (B_j-B_{best}) 
\end{aligned}
\end{equation}
where $stp$ represents a random step and is generated by a symmetric levy distribution \cite{gandomi2013cuckoo}, as: 
\begin{equation}
\small
\begin{aligned}
\label{equation: 17}
stp = \frac{u}{\lvert Y \rvert^{\frac{1}{\beta}}}
\end{aligned}
\end{equation}
where, generally, $\beta=1.5$, and by taking advantage of the $\Gamma(\cdot)$ distribution, we can say:
\begin{equation}
\small
\begin{aligned}
\label{equation: 18}
u = \ell \cdot (\frac{\Gamma(\beta+1) \cdot \sin\left(\frac{\pi\beta}{2}\right)}{\Gamma\left(\frac{\beta+1}{2}\right) \cdot \beta \cdot 2^{(\beta-1)/2}})^{\frac{1}{\beta}}, \hspace{0.2cm} Y=\ell
\end{aligned}
\end{equation}
The Lévy flight process represents a prominent feature of a cuckoo search, facilitating the generation of new candidate solutions or eggs through a random walk \cite{kumar2014ann, yang2010engineering}. Then, current nests $B_j$ are replaced with the new generation of nests $B_{new}$ if they have a better objective function value. Otherwise, they do not change. Next, the available host nests are limited in number, and a host bird can detect the worst foreign eggs with a probability $(P_a\%)$ and should be updated with new ones: 
\begin{equation}
\small
\begin{aligned}
\label{equation: 19}
B_{j(new)}=B_j+\text{rand}(0,1) \cdot (B_{\iota}-B_{\kappa})
\end{aligned}
\end{equation}
if the new candidate has better fitness, as the final step of the first iteration. $\iota$ and $\kappa \in j$ are random integers from 1 to $\eta$. In summary, the steps involved in designing the optimal solution in this paper are outlined as Algorithm 2.
\begin{table}[h!]
\scriptsize
  \centering
  \renewcommand{\arraystretch}{0.4} 
  \colorbox{white!10}{\fbox{%
    \begin{tabular}{p{0.85\linewidth}} 
      \multicolumn{1}{c}{\textbf{Algorithm 2.} Cuckoo search optimization} \\
      \\
      \hspace{0.1cm}\textbf{Output}: $B_{j}$ and $B_{best}\in \mathbb{R}^{1 \times {\zeta}}$.
      \\
      \\
      {\small 1}\hspace{0.2cm} Define an objective function as \cite{bingul2018novel};\\
      {\small 2}\hspace{0.2cm} Initialize candidates $B_j \in \mathbb{R}^{1 \times {\zeta}}$ for $j=1,...,\eta$;\\
      {\small 3}\hspace{0.2cm} Evaluate the objective function value for each $B_j$;\\
      {\small 4}\hspace{0.2cm} \textbf{While} iteration $\leq n_{iteration}$:\\
      {\small 5}\hspace{0.7cm}Find the best candidate $B_{best}$;\\
      {\small 6}\hspace{0.7cm}Generate new candidates $B_{j(new)}$ using Eq. \eqref{equation: 15};\\
      {\small 7}\hspace{0.7cm}Find the objective function value for each $B_{j(new)}$;\\
      {\small 8}\hspace{0.7cm}Store the better candidates, as $B_j$, between $B_{j(new)}$ and $B_j$;\\
      {\small 9}\hspace{0.7cm}Discard $P_a\%$ of the worst candidates;\\
      {\small 10}\hspace{0.55cm}Update them by new candidates $B_{j(new)}$ using Eq. \eqref{equation: 19};\\
      {\small 11}\hspace{0.55cm}Find the objective function value for each $B_{j(new)}$;\\
      {\small 12}\hspace{0.55cm}Store the better candidates, as $B_j$, between $B_{j(new)}$ and $B_j$;\\
      {\small 13}\hspace{0.2cm}\textbf{end}\\  
      {\small 14}\hspace{0.2cm}Find the best candidate $B_{best}=Best[B_j,\ldots, B_\eta]$;\\ 
      {\small 15}\hspace{0.2cm}\textbf{Display} $B_{best}$\\  
    \end{tabular}%
  }}
\end{table}
\subsection{Adaptive subsystem-based control}
By considering the collision-free velocity-bounded trajectories obtained from Section III and \eqref{equation: 3a}, a new transformation of the tracking model is considered, as follows:
\begin{equation}
\small
\begin{aligned}
\label{equation: 4}
 {\Upsilon_1} = {e}_1, \hspace{+0.1cm}
 {\Upsilon_2} = {e}_2 - {\tau}_{0}
\end{aligned}
\end{equation}
The virtual control term, denoted by ${\tau}_0 \in \mathbb{R}^n$, is defined as follows:
\begin{equation}
\small
\begin{aligned}
\label{equation: 5}
{{\tau}_{0}} = - \frac{1}{2} a_{0}{\Upsilon_1}
\end{aligned}
\end{equation}
$a_0$ is a positive constant. The derivative of \eqref{equation: 4} is, as:
\begin{equation}
\small
\begin{aligned}
\label{equation: 6}
\dot{{\Upsilon}}_1= {{\Upsilon}_2}+{\tau}_0, \hspace{+0.2cm}
\dot{{\Upsilon}}_2= {M^{-1}} {\tau}+{F}^*+ {\tau}_d-{\dot{x}}_{2d}\\
\end{aligned}
\end{equation}
The derivative of $\tau_0$ is considered as a term of uncertainty:
\begin{equation}\label{equation:7}
\small
\begin{aligned}
    F^* &= F(x_1, x_2) - \frac{\partial \tau_{0}}{\partial \Upsilon_{1}} \frac{\mathrm{d} \Upsilon_{1}}{\mathrm{d} t}
\end{aligned}
\end{equation}
where ${F}^*=[{F}^*(1),\ldots,{F}^*(n)]$. We can presume the function ${\tau}_{0}$ is bounded and differentiable.
\\
\indent \textbf{Assumption 1}: Because we can practically assume that uncertainties and acceleration are bounded, we assume for all joints ($i=1,\ldots,n$), there are positive constants $m_i$, $\tau_{id}$, and $A_{\text{max}_i} \in \mathbb{R}^+$ such that
\begin{equation}
\small
\begin{aligned}
\label{equation: 881}
&|{{F}}^*(i)| \leq m_i, \hspace{0.1cm} |\tau_{d}(i)| \leq \tau_{id}, \hspace{0.1cm} |{\dot{x}}_{2d}(i)| \leq A_{\text{max}_i}
\end{aligned}
\end{equation}
Thus, following \eqref{equation: 881}, we can find positive constants $m$, $\tau_{\text{max}}$, and $A_{\text{max}}  \in \mathbb{R}^+$, along with a continuously bounded function $H: \mathbb{R}^n \rightarrow \mathbb{R}^+$ with strictly positive values to satisfy the following condition:
\begin{equation}
\small
\begin{aligned}
\label{equation: 8}
&\|{{F}}^*\| \hspace{0.1cm} \leq m H \hspace{0.1cm}, \hspace{0.2cm} \|{\tau}_d\| \hspace{0.1cm} \leq  \tau_{max}, \hspace{0.2cm}\|{\dot{x}}_{2d}\| \hspace{0.1cm} \leq A_{max}
\end{aligned}
\end{equation}
where $||\cdot||$ is the Euclidean norm.\\ 
\indent \textbf{Definition 1} \cite{corless1993bounded,heydari2024robust}: For $t \geq t_0$, we can say the system state $x=[x_1,x_2]^{\top}$ tracks the reference state ${x_d=[{x}_{1d},{x}_{2d}]^{\top}}$ with uniformly exponential convergence if the following condition is satisfied:
\begin{equation}
\small
\begin{aligned}
\label{equation: 9}
\|x(t)- {{x}}_{d}(t)\| \leq \bar{\alpha} e^{-\alpha (t-t_0)} \|x(t_0)-x_d(t_0)\| + \alpha^*
\end{aligned}
\end{equation}
where $\bar{\alpha}$, $\alpha^*$, and  $\alpha \in \mathbb{R}^+$ are positive constants. More precisely, the vector of tracking error ${e}=[e_1, e_2]^{\top}=x-x_d$ is uniformly exponentially stabilized within a defined region $\Psi\left(\tau\right)$ with radius ${R}$, as follows:
\begin{equation}
\small
\begin{aligned}
\label{equation: 10}
&\Psi\left({R}\right):=\left\{{e} \mid\|{e} \| \leq {R} = \alpha^*\right\}
\end{aligned}
\end{equation}
The actual torque control ${\tau}$ is proposed, as follows:
\begin{equation}
\small
\begin{aligned}
\label{equation: 11}
{\tau}=\frac{-1}{2}{M}(a_{1}+b_{1}\hat{\rho}){{\Upsilon}_2}-{M}{\Upsilon_1}
\end{aligned}
\end{equation}
$a_1 $ and $b_{1}$ are positive constants, and $\hat{\rho}$ is the adaptive law:
\begin{equation}
\small
\begin{aligned}
\label{equation: 12}
\dot{\hat{\rho}} &= -{c_1}{r_1}\hat{\rho}+\frac{1}{2}b_{1}{c_1}\|{\Upsilon_2}\|^{2}\\
\end{aligned}
\end{equation}
where $c_1$, $c_2 $, and $r_{1}$ are positive constants.\\
\indent \textbf{Assumption 2}\cite{heydari2024robust}: 
According to the general solution of the given linear first-order ordinary differential equation in \eqref{equation: 12}, by choosing an initial condition $\hat{\rho}(0)\geq 0$, we can claim that for all $t \geq 0$, $\hat{\rho}\geq 0$.\\
\indent By assuming the vector of the adaptive law error is $\tilde{\rho}=\hat{\rho}-\rho^*$, where $\rho^*$ is an unknown and constant parameter:
\begin{equation}
\small
\begin{aligned}
\label{equation: 13}
&\dot{\tilde{\rho}}=-{c_1}{r_1}\tilde{\rho}+\frac{1}{2}b_{1}{c_1}\|{{\Upsilon}_2}\|^2-{c_1}{r_1}{\rho}^*\\
\end{aligned}
\end{equation} 
where $\rho^*$ is defined, as follows:
\begin{equation}
\small
\begin{aligned}
\label{equation: 14}
 \hspace{-0.1cm}  \rho^*=& b_{1}^{-1}(\mu m^2+\nu_{1}{\tau_{max}^2}+\nu_{{2}}A_{max}^2)
\end{aligned}
\end{equation}
where $\mu$, $\nu_1$, and $\nu_2$ are unknown positive constant. The steps for designing the subsystem-based controller structure in this paper are outlined in Algorithm 3 and Figure 2, respectively.
\vspace{-0.5cm}
\begin{table}[h!]
\scriptsize
  \centering
  \renewcommand{\arraystretch}{0.4} 
  \colorbox{white!10}{\fbox{%
    \begin{tabular}{p{0.85\linewidth}} 
      \multicolumn{1}{c}{\textbf{Algorithm 3.} Auto-tuned subsystem-based adaptive control strategy} \\
      \\
      \hspace{0.4cm}\textbf{Input}: robot states from sensors, $a_t={x_{2d}}$ (\textbf{algorithm 1}), \\
      \hspace{1.1cm}$[a_0, a_1, b_1, c_1, r_1]$ from \textbf{algorithm 2}.\\
      \hspace{0.4cm}\textbf{Output}: control input ${\tau}$.\\
      \\
      {\small 1}\hspace{0.5cm}\textbf{While new action is provided:} (from step 8 of \textbf{algorithm 1})\\
      {\small 2}\hspace{1.1cm}${e_1}={x_1}-{x_{1d}}$;\\
      {\small 3}\hspace{1.1cm}${e_2}={x_2}-{{x}_{2d}}$;\\
      {\small 4}\hspace{1.1cm}${\Upsilon_1} = {e}_1$;\\
      {\small 5}\hspace{1.1cm}${{\tau}_{0}} = - \frac{1}{2} a_{0}{\Upsilon_1}$;\\
      {\small 6}\hspace{1.1cm}${\Upsilon_2} = {e}_2 - {\tau}_{0}$;\\
      {\small 7}\hspace{1.1cm}$\dot{\hat{\rho}}= -{c_1}{r_1}\hat{\rho}+\frac{1}{2}b_{1}{c_1}\|{\Upsilon_2}\|^{2}$;\\
      {\small 8}\hspace{1.1cm}${\tau}=-\frac{1}{2} M(a_{1}+b_{1}\hat{\rho}){{\Upsilon}_2}-M{\Upsilon_1}$;\\
      {\small 9}\hspace{1.0cm}\textbf{Display} $\tau$;\\
      {\small 10}\hspace{0.5cm}\textbf{end}\\
    \end{tabular}%
  }}
\end{table}
\\
\indent \textbf{Remark 1}: All three algorithms provided (learning, CSO optimization, and control phase) operate actively within the environment. Specifically, the low-level control gains $[a_0, a_1, b_1, c_1, r_1]$ are determined as $B_j$ or $B_j(new)$ from Algorithm 2 while the iteration count is less than or equal to $n_{\text{iteration}}$. After the tuning process is completed, they will represent $B_{\text{best}}$ for the remaining control tasks. Furthermore, the learning action (reference velocities of joints) $a_t$ from Algorithm 1 is utilized in Algorithm 3 at each step.
\begin{figure*} [t]
    \centering
    \includegraphics[width=0.85\textwidth, height=5.8cm]{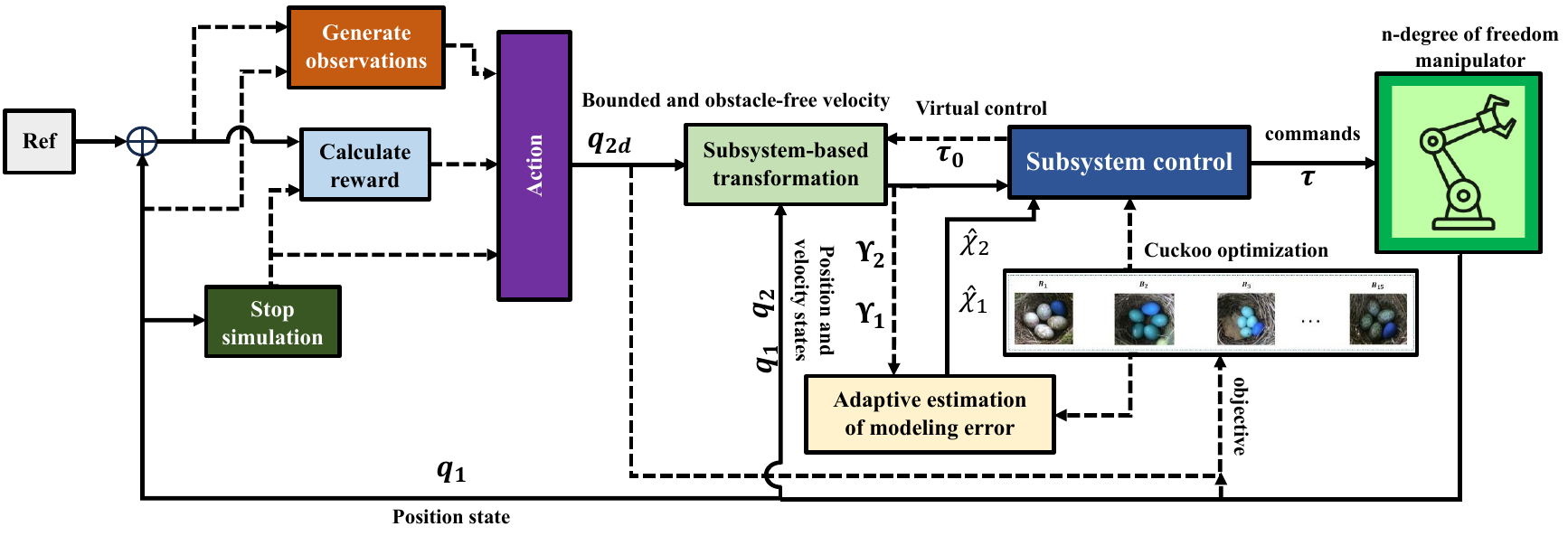}
    \caption{Integration of the low-level robust subsystem-based control with collision-free DRL-based motion planner.}
    \label{fig11}
\end{figure*}
\subsection{Stability analysis}
\textbf{Theorem 1}: Consider the manipulator system \eqref{equation: 2}, the control strategy \eqref{equation: 11} along with the adaptive law in \eqref{equation: 12}. The tracking errors between the actual states and the reference trajectories generated by the DRL-based motion planner are uniformly exponentially stable.\\
\indent \textbf{Proof}: A Lyapunov function is suggested as follows:
\begin{equation}
\small
\begin{aligned}
\label{equation: 20}
&V_{1} =\frac{1}{2} \hspace{0.1cm} [ {\Upsilon}_1^{\top}{\Upsilon}_1]
\end{aligned}
\end{equation} 
After differentiating $V_1$ and inserting \eqref{equation: 6}, we obtain:
\begin{equation}
\small
\begin{aligned}
\label{equation: 21}
\dot{V}_{1}=   {\Upsilon}_1^{\top} [{\Upsilon}_2+{\tau}_{0}]={\Upsilon}_1^{\top} {\Upsilon}_2+ {\Upsilon}_1^{\top} {\tau}_{0}
\end{aligned}
\end{equation} 
By considering the description  of \eqref{equation: 5}, we obtain:
\begin{equation}
\small
\begin{aligned}
\label{equation: 24}
\dot{V}_1\leq& \hspace{0.1cm} {\Upsilon}_1^{\top} {\Upsilon}_2  - \frac{1}{2} a_{0} ||{\Upsilon}_1||^2
\end{aligned}
\end{equation} 
considering \eqref{equation: 20}:
\begin{equation}
\small
\begin{aligned}
\label{equation: 25}
\dot{V_{1}}\leq& -\beta_{1} V_{1} + {\Upsilon}_1^{\top} {\Upsilon}_2
\end{aligned}
\end{equation} 
where $\beta_1 = {a_0}$.
We have another Lyapunov function, as:
\begin{equation}
\small
\begin{aligned}
\label{equation: 27}
&V_{2} =\frac{1}{2} \hspace{0.1cm} [ {\Upsilon}_2^{\top}{\Upsilon}_2+{c^{-1}_{1}} \tilde{\rho}^{2} ]
\end{aligned}
\end{equation} 
\vspace{-0.1cm}
By differentiating $V_2$ and inserting \eqref{equation: 6} into \eqref{equation: 27}, we have:
\begin{equation}
\small
\begin{aligned}
\label{equation: 28}
\dot{V}_2=&{\Upsilon_2}^{\top} [{M^{-1}} {\tau}+{F}^*+ {\tau}_d-{\dot{x}}_{2d}]+c_1^{-1} \tilde{\rho}\dot{\tilde{\rho}}
\end{aligned}
\end{equation}
Considering positive constants $\mu$, $\nu_1$, and $\nu_2$ and \eqref{equation: 8}:
\begin{equation}
\small
\begin{aligned}
\label{equation: 29}
\dot{V}_2 \leq &{\Upsilon_2}^{\top} {M^{-1}} {\tau}+||{\Upsilon_2}||mH+ ||{\Upsilon_2}|| {\tau}_{max}+||{\Upsilon_2}|| A_{max}\\
&+c_1^{-1} {\tilde{\rho}}{\dot{\tilde{\rho}}}
\end{aligned}
\end{equation}
Following Young's inequality: 
\begin{equation}
\small
\begin{aligned}
\label{equation: 30}
\hspace{-0.4cm}\dot{V}_2\leq& {\Upsilon_2}^{\top} {M^{-1}} {\tau}+ \frac{1}{2} ||{\Upsilon_2}||^2 \mu m ^2+ \frac{1}{2} || {\Upsilon_2} || ^2 \nu_1 {\tau_{max}}^2\\
&+\frac{1}{2}\nu_2^{-1}+ \frac{1}{2} \mu^{-1} H^2+ \frac{1}{2} \nu_1^{-1}+\frac{1}{2}\nu_2 A_{max}^2 ||{\Upsilon_2}||^2\\
&+c_1^{-1} {\tilde{\rho}}{\dot{\tilde{\rho}}}
\end{aligned}
\end{equation} 
By considering \eqref{equation: 11} and \eqref{equation: 13}:
\begin{equation}
\small
\begin{aligned}
\label{equation: 32}
\hspace{-0.2cm}\dot{V}_2&\leq -\frac{1}{2}a_{1}||{{\Upsilon}_2}||^2-\frac{1}{2}b_{1}\hat{\rho}||{{\Upsilon}_2}||^2-{\Upsilon}_2^{\top}{\Upsilon}_2+ \frac{1}{2} \nu_1^{-1}\\
&+ \frac{1}{2} ||{\Upsilon_2}||^2 \mu m ^2+ \frac{1}{2} || {\Upsilon_2} || ^2 \nu_1 {\tau_{max}}^2+\frac{1}{2}\nu_2 A_{max}^2 ||{\Upsilon_2}||^2\\
&+ \frac{1}{2} \mu^{-1} H^2+\frac{1}{2}\nu_2^{-1}-{r_1}\tilde{\rho}^2+\frac{1}{2}b_{1}\|{{\Upsilon}_2}\|^2\tilde{\rho}-{r_1}{\rho}^*\tilde{\rho}
\end{aligned}
\end{equation} 
Regarding \eqref{equation: 14}, and knowing that $\tilde{\rho}=\hat{\rho}-\rho^*$:
\begin{equation}
\small
\begin{aligned}
\label{equation: 33}
\hspace{-0.2cm}\dot{V}_2\leq& -\frac{1}{2}a_{1}||{{\Upsilon}_2}||^2-{\Upsilon}_2^{\top}{\Upsilon}_2+ \frac{1}{2} \mu^{-1} H^2+ \frac{1}{2} \nu_1^{-1}+\frac{1}{2}\nu_2^{-1}\\
&-{r_1}\tilde{\rho}^2-{r_1}{\rho}^*\tilde{\rho}
\end{aligned}
\end{equation} 
After dividing ${}{r_1}{\tilde{\rho}}^2$ into $\frac{1}{2}{}{r_1}{\tilde{\rho}}^2+\frac{1}{2}{}{r_1}{\tilde{\rho}}^2$, we can obtain:
\begin{equation}
\small
\begin{aligned}
\label{equation: 34}
\dot{V}_2\leq & -\frac{1}{2}a_{1}||{{\Upsilon}_2}||^2-{\Upsilon}_2^{\top}{\Upsilon}_2+ \frac{1}{2} \mu^{-1} H^2+ \frac{1}{2} \nu_1^{-1}\\
&+\frac{1}{2}\nu_2^{-1}-\frac{1}{2}{r_1}{\tilde{\rho}}^2+\frac{1}{2}{}{r_1}{\rho^*}^2 
\end{aligned}
\end{equation} 
By defining $\beta^*= \frac{1}{2} \sum_{i=1}^{2} \nu_i^{-1}+\frac{1}{2}{r_1}{\rho^*}^2$:
\begin{equation}
\small
\begin{aligned}
\label{equation: 340}
\dot{V}_2 \leq -\beta_2 V_2 -{\Upsilon}_2^{\top}{\Upsilon}_2 + \frac{1}{2} \mu^{-1} H^2 + \beta^*
\end{aligned}
\end{equation} 
where $\beta_2 = \min [{a_1},\hspace{0.3cm}{c_1}r_1]$.
Now, we can establish a Lyapunov function for the entire system, by considering the Lyapunov functions \eqref{equation: 20} and \eqref{equation: 27}, as follows:
\begin{equation}
\small
\begin{aligned}
\label{equation: 36}
&V =V_1+V_2
\end{aligned}
\end{equation}
After the derivative, and after inserting \eqref{equation: 25}, and \eqref{equation: 34}:
\begin{equation}
\small
\begin{aligned}
\label{equation: 37}
\dot{V} \leq& -\sum_{i=1}^2 \beta_i V_i + \frac{1}{2} \mu^{-1} H^2+ \beta^*
\end{aligned}
\end{equation}
To simplify \eqref{equation: 37}, we can say:
\begin{equation}
\small
\begin{aligned}
\label{equation: 38}
\dot{V}\leq& -\beta_{min} V+ \frac{1}{2} \mu^{-1} H^2 + \beta^*
\end{aligned}
\end{equation}
where $\beta_{min} = \min [\beta_1,\hspace{0.1cm}\beta_2]$.
Thus, regarding \eqref{equation: 38}, based on the Definition 1, Theorem 1 is proved.

\begin{table}[h]
  \captionsetup{position=top}
  \caption{DRL environment parameters.}
  \centering
  \scriptsize
  \begin{tabular}{lc}
    \toprule
    {\textbf{Parameters}} & {\textbf{Value}} \\
    \midrule
    Reach Reward & \bm{$200$} \\
    Boundary Penalty & \bm{$50$} \\
    Collision Penalty & \bm{$100$} \\
    Joint Limits (rad) & \bm{$x_1(1): [-\pi/4, 3\pi/4]$}, \bm{$x_1(2): [-5\pi/6, 5\pi/6]$} \\
    Reach Threshold (m) & \bm{$0.04$}  \\
    \bottomrule
  \end{tabular}
\end{table}
\section{Numerical Evaluation}
We applied the proposed control strategy to the 2-DoF vertical plane robot expressed in \cite{humaloja2021decentralized} with link lengths $1$m and $0.8$m. Both the auto-tuned low-level control and the manipulator model collectively form our environment. This implies the actions generated by the agent enter a closed loop encompassing both. Environment parameters are configured as in Table I. Simulation resets and the beginning of a new episode occurs when a joint surpasses its limits, the manipulator attains the goal, a collision occurs, or the number of simulation steps exceeds the limit. The SAC parameters are configured as in Table II. The policy runs at a 33.33-Hz frequency and the controller runs at 1 KHz in the simulation. Unknown frictions and external load effects are expressed as:\\
\begin{equation}
\small
\begin{aligned}
\label{equation: 40}
&{F}=\left[\begin{array}{l}
0.5 \cos \left(0.7{x}_2(2)\right) \\
-1.1 \cos \left(1.8 x_1(2)\right)+1.8 \cos \left(0.3{x}_1(2)\right) \\
\end{array}\right]\\
&{\tau_d}=\left[\begin{array}{l}
3 \cos(2t) \\
-0.2 \hspace{0.1cm} \text{rand}(0,1) \\
\end{array}\right]
\end{aligned}
\end{equation}
The parameters associated with CSO are considered as follows: $\zeta = 8$, $\beta = 1.5$, $n_{\text{iteration}} = 200$, $P_a=25\%$, and $\eta = 15$.
This simulation considers five different tip obstacles, the same initial conditions, and four random targets in task space, as shown in Fig. 3. The control process ceased when the error between the tip of the robot and the target met the condition of having an error of less than 4 cm, as shown in Fig 4. The figure illustrates the target error from the initial to the target position in task space for four tasks. Although target (d) is the second farthest, the task's goal was achieved the soonest in contrast to the task (b) pattern, suggesting that the path involved lower complexity and fewer obstacles, leading to a quicker accomplishment. In summary, the performance of the proposed control strategy is indicated in Table III. It shows that the robot system took the least amount of time (about $5.5$sec) to reach the target in task (d), while task (c) required the most time (more than $11$sec).  
\begin{table}[h]
  \captionsetup{position=top}
  \caption{DRL agent parameters.}
  \centering
  \scriptsize
  \begin{tabular}{lc}
    \toprule
    {\textbf{Parameters}} & {\textbf{Value}} \\
    \midrule
    Velocity Bounds (rad/s) & \bm{$x_2(1):[-0.1,+0.1]$}, \bm{$x_2(2):[-0.4,+0.4]$} \\
    Batch size & \bm{$512$} \\
    Target smoothing factor & \bm{$0.001$} \\
    Experience buffer length & \bm{$1000000$} \\
    Initial random steps & \bm{$1000$} \\
    Discount factor & \bm{$0.995$} \\
    Training episodes & \bm{$20000$} \\
    Time step & \bm{$0.03$} \\
    Max time steps/episode & \bm{$1000$} \\
    Learning rate & \bm{$0.001$} \\
    \bottomrule
  \end{tabular}
\end{table}

\begin{figure}[h!] 
    \centering
    \scalebox{1.00}{\includegraphics[trim={0cm 0.0cm 0.0cm 0cm},clip,width=\columnwidth]{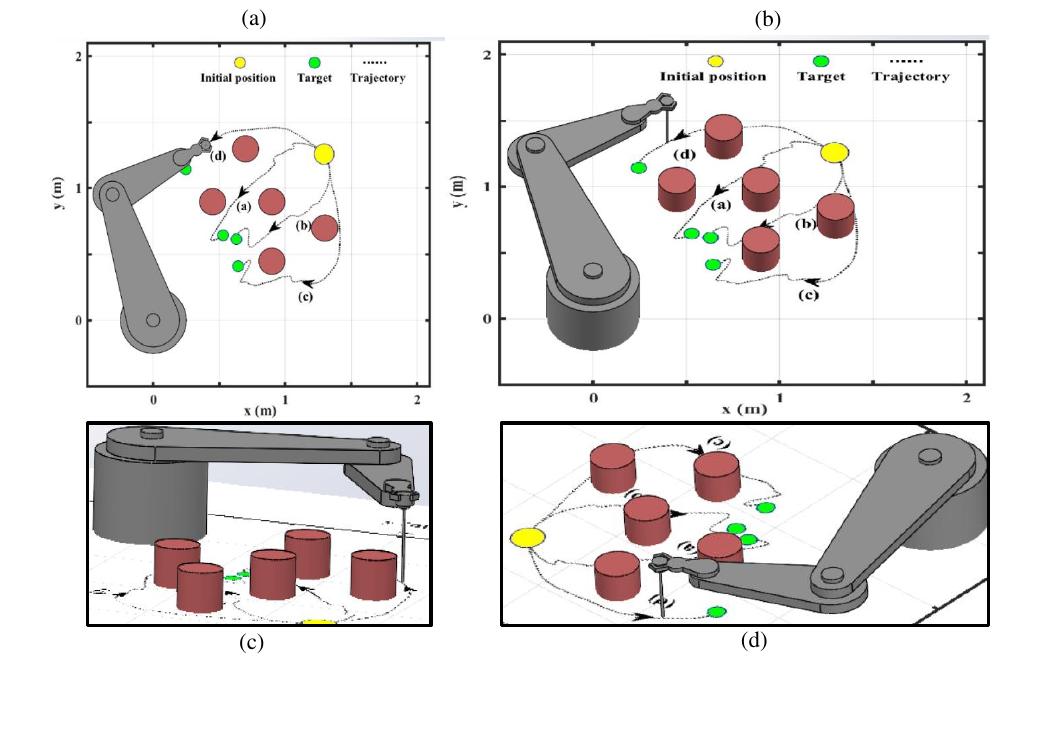}}
    \caption{{Manipulator's tip tracking collision-free trajectories for non-repetitive tasks in different views.}}
    \label{fig33}
\end{figure}
In addition, tasks (a) and (c) experienced the highest torque amplitudes ($400$Nm), while task (d) had the least torque effort ($220$Nm) in reaching the target, as shown in Fig. 5. The figure illustrates that tasks involve more changes in direction necessitating higher torque generation. 

\begin{table}[h]
  \captionsetup{position=top}
  \caption{Performance of the proposed control.}
  \centering
  \tiny
  \begin{tabular}{cccccc}
    \toprule
    {\textbf{Goal-reaching}} & {\textbf{Target}} &
    {\textbf{Target}} &
    {\textbf{tracking}} & {\textbf{Max. torque}} &
    {\textbf{optimization}} \\
    {\textbf{tasks}}& 
    {\textbf{time (s)}}&
    {\textbf{error (cm)}}& 
    {\textbf{error (cm)}}&
    {\textbf{amplitude (N.m)}}&
    {\textbf{time (s)}}\\
    \midrule
    \textbf{(a)} & \bm{$9.176$} & \bm{$0.025$} & \bm{$0.0006$} & \bm{$400$} & \bm{$0.01$}\\
    \textbf{(b)} & \bm{$9.92$} & \bm{$0.037$} & \bm{$0.0008$} & \bm{$360$} & \bm{$0.03$}\\
    \textbf{(c)} & \bm{$11.16$} & \bm{$0.029$} & \bm{$0.001$} & \bm{$400$} & \bm{$0.05$} \\
    \textbf{(d)} & \bm{$5.456$} & \bm{$0.039$} & \bm{$0.0002$} & \bm{$220$} & \bm{$0.009$} \\
    \bottomrule
  \end{tabular}
\end{table}

By utilizing CSO, the controller gains converged to the following values: $a_0 = 668$, $a_1 = 552$, $b_1 = 1.8$, $c_1 = 0.001$, and $r_1 = 0.69$. The performance of the tracking position references by two joints is depicted in Fig. 6, where all states converge rapidly to the respective references before 0.05 seconds when control gains are optimized. The proposed low-level control was not only auto-tuned and lacks information about the environment but is also more robust, and capable of compensating for uncertainties in \eqref{equation: 40}.

\begin{figure}[h!]
    \centering
    \scalebox{0.8}{\includegraphics[trim={0cm 0.0cm 0.0cm 0cm},clip,width=\columnwidth]{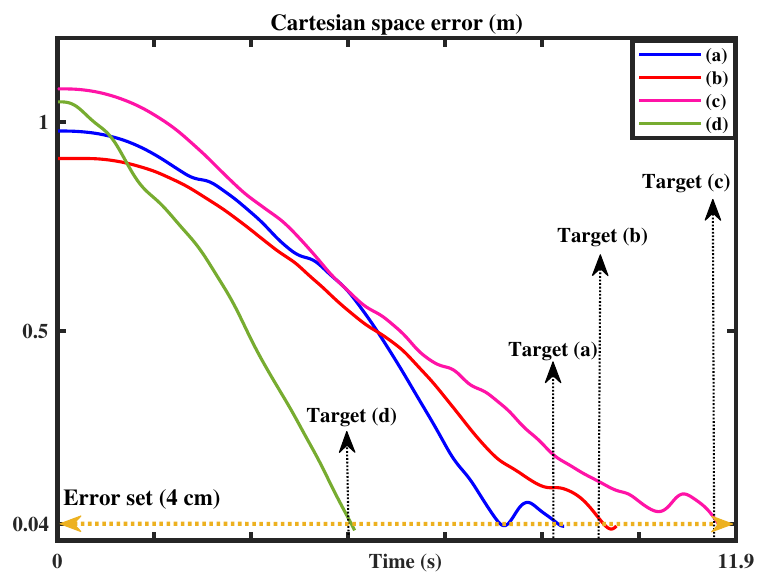}}
    \caption{The position target error in task space.}
    \label{fig33}
\end{figure}

\begin{figure}[h!]
    \centering
    \scalebox{0.8}{\includegraphics[trim={0cm 0.0cm 0.0cm 0cm},clip,width=\columnwidth]{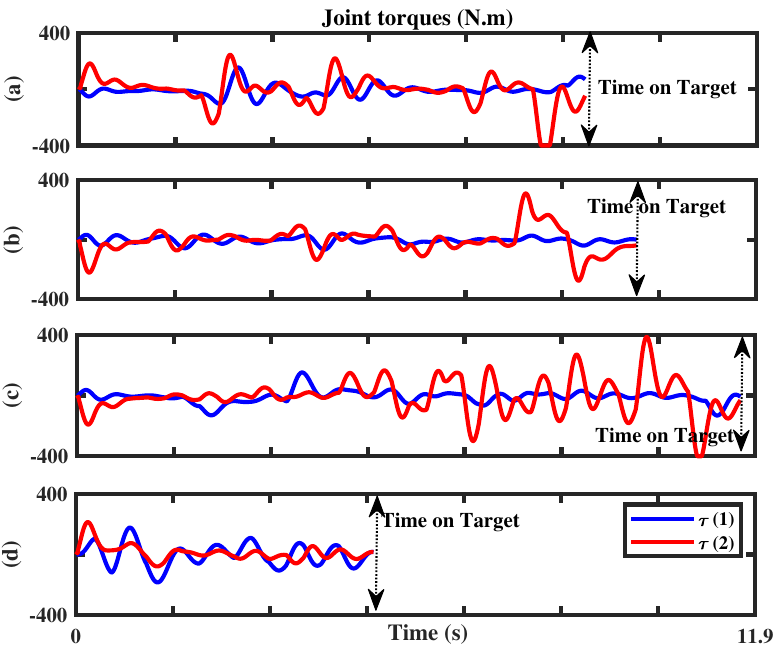}}
    \caption{Torques generated by joints to perform control targets.}
    \label{fig33}
\end{figure}
\begin{figure}[h!]
    \centering
    \scalebox{0.85}{\includegraphics[trim={0cm 0.0cm 0.0cm 0cm},clip,width=\columnwidth]{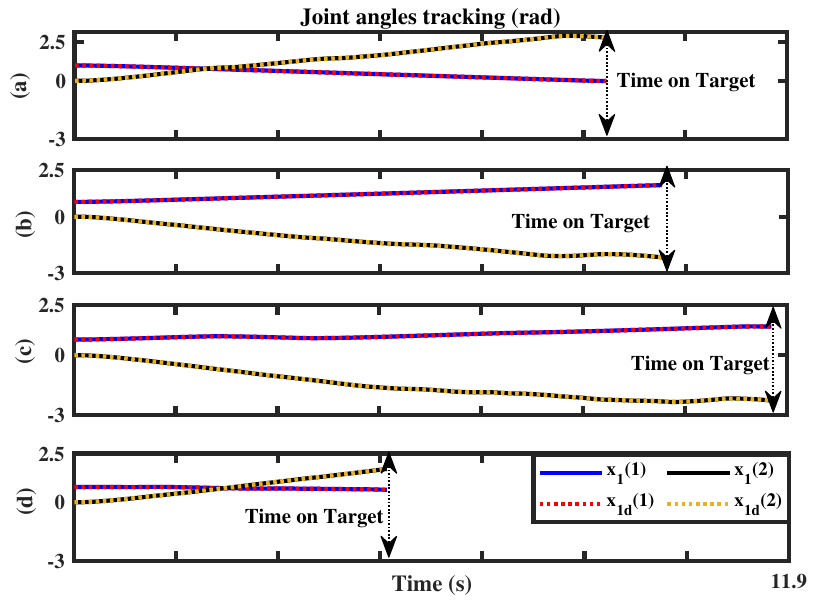}}
    \caption{Position tracking of the DRL-generated.}
    \label{fig33}
\end{figure}

Our investigation, comparing the proposed control with PID control as the low-level control in the same condition, validated improvements in speed convergence and error accuracy, achieving approximately $10\%$ faster convergence and $18\%$ greater accuracy. Therefore, we believe that utilizing advanced control methods, rather than classical ones, as a low-level control integrated with a DRL-based planner, often leads to superior performance in stability, transient response, steady-state error, and robustness, especially for collision-free reaching tasks.

\section{Conclusion}
To enhance the performance of goal-reaching tasks for an n-DoF manipulator, this paper proposed a novel, robust low-level control integrated with a velocity-bounded, collision-free, and non-repetitive DRL-based motion planner. Hence, the SAC algorithm was employed to generate motion trajectories for goal-reaching tasks in the presence of obstacles for random target points, feeding a low-level robust adaptive strategy to generate sufficient joint torques, ensuring exponential convergence of tracking errors in the presence of unknown uncertainties, and external disturbances. In addition, gains were optimized using a CSO tailored to minimize the rise time, settling time, maximum overshoot, and steady-state tracking error. The framework operated dynamically within the environment, with CSO dictating low-level control parameters while DRL actions were concurrently utilized in the control section.

\bibliographystyle{IEEEtran}
\bibliography{ref}

\end{document}